\documentclass[runningheads]{llncs}
\usepackage{graphicx}
\usepackage{comment}
\usepackage{amsmath,amssymb} 
\usepackage{color}
\usepackage{multirow}
\usepackage{listings}

\usepackage{times}
\usepackage{epsfig}
\usepackage{pifont}
\usepackage{cite}
\usepackage{appendix}

\definecolor{dkgreen}{rgb}{0,0.6,0}
\definecolor{gray}{rgb}{0.5,0.5,0.5}
\definecolor{mauve}{rgb}{0.58,0,0.82}

\begin{document}
\pagestyle{headings}
\mainmatter
\def\ECCVSubNumber{547}  

\title{Delving into the Imbalance of Positive Proposals in Two-stage Object Detection} 

\author{Zheng Ge\inst{1,2} \and
Zequn Jie\inst{2} \and
Xin Huang\inst{1,2} \and 
Chengzheng Li\inst{3} \and
Osamu Yoshie\inst{1}}

\institute{Waseda University \and Tencent AI Lab \and Nanjing University of Science and Technology \\
\email{jokerzz@fuji.waseda.jp;zequn.nus@gmail.com;koushin@toki.waseda.jp;\\
czhengli@njust.edu.cn;yoshie@waseda.jp}\\
}

\maketitle

\begin{abstract}
Imbalance issue is a major yet unsolved bottleneck for the current object detection models. In this work, we observe two crucial yet never discussed imbalance issues. The first imbalance lies in the large number of low-quality RPN proposals, which makes the R-CNN module (\emph{i.e.}, post-classification layers) become highly biased towards the negative proposals in the early training stage. The second imbalance stems from the unbalanced ground-truth numbers across different testing images, resulting in the imbalance of the number of potentially existing positive proposals in testing phase. To tackle these two imbalance issues, we incorporates two innovations into Faster R-CNN: 1) an R-CNN Gradient Annealing (RGA) strategy to enhance the impact of positive proposals in the early training stage. 2) a set of Parallel R-CNN Modules (PRM) with different positive/negative sampling ratios during training on one same backbone. Our RGA and PRM can totally bring \textbf{2.0\%} improvements on AP on COCO\cite{lin2014microsoft} $minival$. Experiments on CrowdHuman\cite{shao2018crowdhuman} further validates the effectiveness of our innovations across various kinds of object detection tasks.
\end{abstract}

\begin{figure}[t]
\begin{center}
\includegraphics[width=0.7\linewidth]{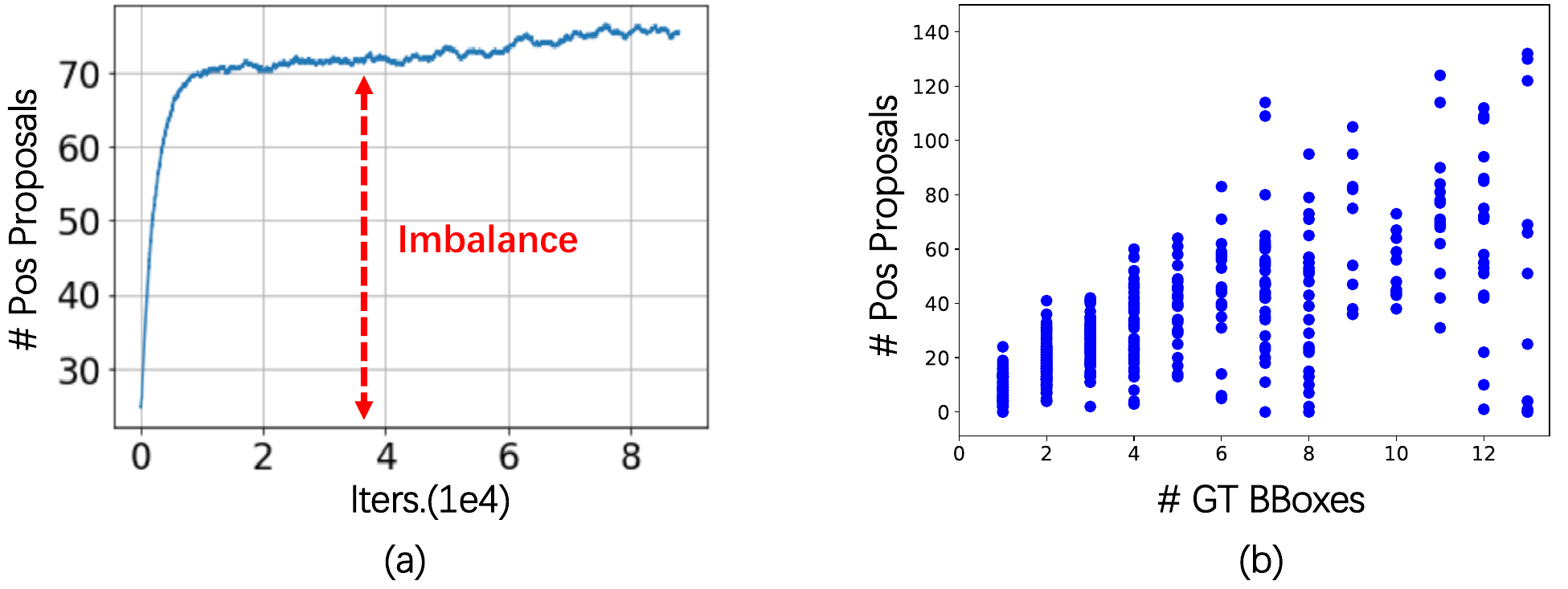}
\end{center}
   \vspace{-0.5cm}
   \caption{(a). The actual number of sampled positive proposals during a training process of Faster R-CNN. (b) shows the distributions of positive proposal numbers generated by a well-trained RPN model w.r.t. different numbers of ground-truth instances in testing images.}
   \vspace{-0.2cm}
   \label{fstart}
\end{figure}

\section{Introduction} \label{sec:intro}

In recent years, the great success of deep learning pushes forward the state-of-the-art object detection approaches, \emph{e.g.}, Faster R-CNN~\cite{ren2015faster}, SSD~\cite{liu2016ssd} and Cascade R-CNN~\cite{cai2018cascade}. However, most of existing works focus on the novel detection pipeline (\emph{i.e.}, one-stage and two-stage detectors) and network architecture design (\emph{e.g.}, Feature Pyramid Network~\cite{lin2017feature} and Path Aggregation Network~\cite{liu2018path}), less attention is paid to the training paradigm. 

Imbalance is a severe issue when training an object detector. A few existing works notice and try to address the imbalance issue including OHEM~\cite{shrivastava2016training}, RetinaNet~\cite{lin2017focal} and Libra R-CNN~\cite{pang2019libra}. These works mainly deal with the training sample imbalance, \emph{e.g.}, the imbalance between positive/negative samples and difficult/easy samples. However, the imbalance issue lies in far more than the aforementioned situations, and many ignored imbalance issues are preventing the power of well-designed model architectures from being fully exploited.

``Jointly training'' for two-stage detectors has become a mainstream in many popular open-source frameworks (\emph{e.g.}~\cite{chen2019mmdetection,wu2019detectron2}) due to its convenience and desirable performance. In this paper, we investigate two crucial yet never discussed imbalance issues based on such training scheme. The first imbalance comes from the large number of low-quality RPN proposals in the early training stage. Such low-quality RPN proposals can hardly provide a sufficient number of positive samples for training the R-CNN module (\emph{i.e.}, layers from RoI alignment~\cite{he2017mask} to the final classification and bounding box regression), leading to an extremely unbalanced number of positive proposals along whole training process. Fig.~\ref{fstart}(a) shows the actual number of positive proposals used for the R-CNN module \emph{v.s.} training iterations. In the early training stage, the dominant negative proposals would push the model highly biased to the background class side. Although the issue is gradually relieved as the qualities of the RPN proposals increase in training, the significant initial learning bias towards to the background class hurts both training efficiency and model performance which are clearly shown in Sec.~\ref{sec4-1}.
 
\begin{figure}[t]
\begin{center}
\includegraphics[width=0.7\linewidth]{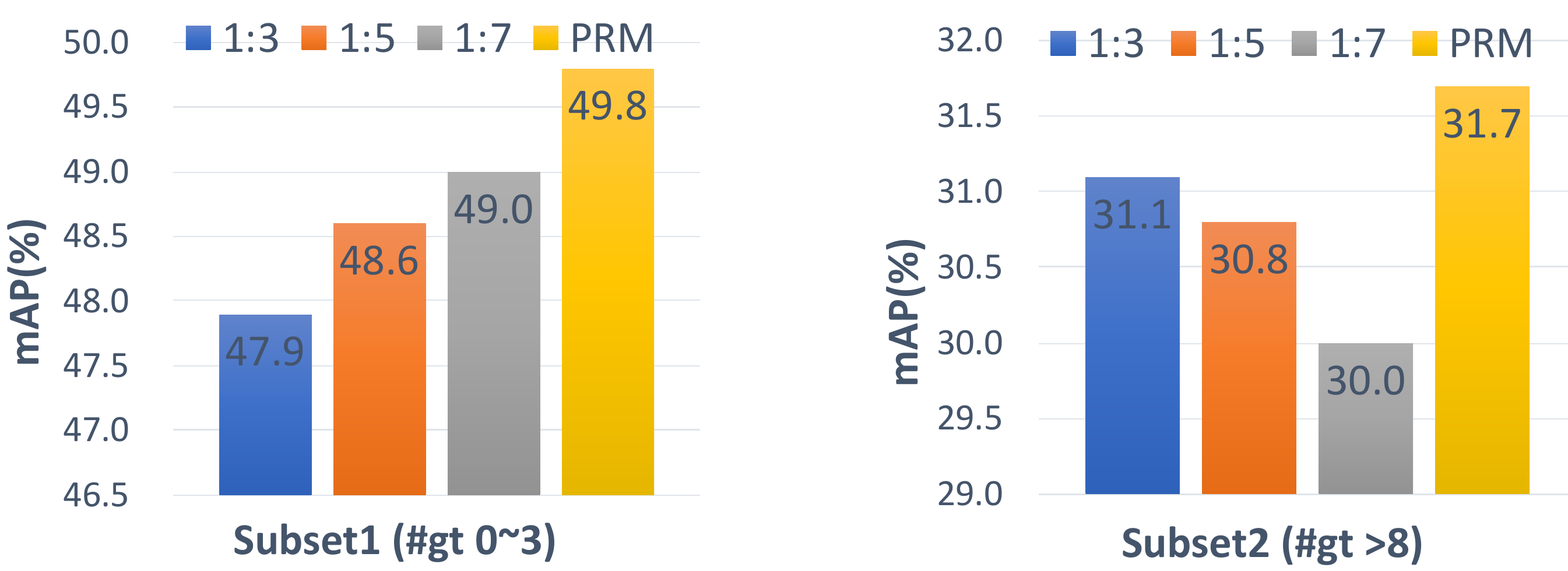}
\end{center}
   \vspace{-0.5cm}
   \caption{Performance of Faster R-CNN w/ and w/o PRM on two subsets with different ground-truth instance numbers of COCO $minival$, trained using different sampling ratios. Faster R-CNN w/ PRM consistently surpasses other counterparts on both subsets.}
   \label{fdivide}
   \vspace{-0.2cm}
\end{figure}

Another serious but also ignored imbalance issue stems from the inconsistency on the potentially existing number of positive proposals across different testing images, which may requires different optimal training strategies (\emph{i.e. different positive/negative sampling ratios for training the R-CNN module}). Fig.~\ref{fstart} (b) shows the distributions of positive proposal numbers generated by a well-trained RPN model w.r.t. different numbers of ground-truth instances in testing images. One can find that for a well-trained RPN model, the more ground-truth instances a testing image contains, the more positive proposals it will probably provide to the following R-CNN module. Notice that here the positive proposals include all the positive ones regarding to all the classes except for the background. As well known, for an image classification task, a better per-image accuracy on the testing set can be achieved when the sample ratios on all the classes of the training set are more consistent with that of the testing set. In object detection, a similar phenomenon is observed. Using various positive/negative proposal sampling ratios during training, the testing performance will be different accordingly. Fig.~\ref{fdivide} \textbf{Left} and \textbf{Right} show the detection performance on two subsets of COCO $minival$ containing images with the ground-truth instance number ranging in [1,3] and [8,+$\infty$) respectively, using different training sampling ratios. One can find that a higher positive sampling ratio is desired for the testing images with larger numbers of ground-truth instances, and vice versa. Such a phenomenon is natural. When setting a high sampling ratio of positive proposals, the model tends to predict higher scores for all proposals, which is beneficial for the cases with larger numbers of ground-truth instances in images. On the contrary, the model will give positive predictions more prudently. Therefore, the inconsistency between a single training sampling ratio and the diverse ground-truth instance numbers in testing images makes the current training sampling strategy a sub-optimal solution. 

To overcome the first observed imbalance issue happened at the early training stage, a novel R-CNN Gradient Annealing (RGA) strategy is proposed. The gradients of positive proposals are magnified to avoid being overwhelmed by the gradients of the huge number of negative proposals at the beginning. As the training progresses, the magnification factor is gradually decreased to guarantee the gradients from both positive and negative proposals can always rival each other.

To address the second imbalance issue, we propose to build two parallel R-CNN modules (PRM) on top of a shared backbone and RPN. To strengthen the adaptation of the detector to a wide range of ground-truth instance numbers contained in testing images, more diverse sampling ratios are simultaneously expected during training. To this end, the two R-CNN modules are trained with two sets of proposals sampled using different positive/negative ratios from a same set of proposals generated by the shared RPN. In the testing phase, proposal-level ``average ensemble'' on the results of the two R-CNN modules can thus be easily performed, effectively incorporating the knowledge learned with diverse class biases. As an extra bonus, without performing ``average ensemble'', PRM also enhances the detection performance of each individual R-CNN module compared to training them solely, as seen in Table~\ref{table1}. This phenomenon reveals that apart from result ensemble, gradient ensemble in the detector backbone from the two R-CNN modules also boosts the model's adaptation to the testing images with significantly diverse ground-truth instance numbers.

\begin{table}[t]
   \caption{The results of the PRM with the sampling ratios of 1:1 and 1:9 respectively. Not only the ensemble result outperforms its baseline by 1.0\%, but also the performance of each single R-CNN module gets improved.}
   \label{table1}
\begin{center}
\begin{tabular}{cc|ccc}
\hline
\multirow{2}{*}{1:1} & \multirow{2}{*}{1:9} & \multicolumn{3}{c}{AP (\%)} \\ \cline{3-5} 
                     &                      & R-CNN1 & R-CNN2 & Ensemble \\ \hline
\checkmark &                      &   36.3    &  -  &   -       \\
           &    \checkmark                  &   -    &  34.9  &   -       \\
\checkmark &    \checkmark       &   \textbf{37.0}    &      \textbf{36.9} & \textbf{37.3} \\ \hline
\end{tabular}
\vspace{-0.5cm}
\end{center}
\end{table}

Experiments on COCO and CrowdHuman benchmarks strongly validate the efficacy of RGA and PRM over several two-stage detector baselines, \emph{e.g.}, Faster R-CNN and Cascade R-CNN.

To summarize, our work has the following contributions:

\begin{enumerate}
\item We observe two crucial but never discussed imbalance issues in object detection and analyze the reasons of the issues, indicating another improvement space of the current two-stages object detection approaches.
\item We propose an R-CNN Gradient Annealing strategy to remedy the lack of positive proposals in the early stage of the training phase. 
\item We propose a novel PRM which integrates a set of parallel R-CNN modules trained by different sampling ratios on one same backbone. Our PRM can alleviate the inconsistency of the number of positive proposals along different testing samples.
\end{enumerate}

\section{Related Work}

\subsection{Deep Architectures for Object Detection}
Recently, deep learning based object detection methods are popularized by two-stage and one-stage detectors. Two-stage detectors generate a set of region proposals, and then refine them by region-wise classification and regression. To reduce redundant computation of feature extraction in R-CNN\cite{girshick2014rich}, \cite{he2015spatial} and \cite{girshick2015fast} propose Spatial-Pyramid-Pooling and RoI-Pooling layers respectively, leading to remarkable improvements of speed and accuracy. After that, Region Proposal Network (RPN) is proposed in Faster R-CNN\cite{ren2015faster} to improve the efficiency of detectors. RPN also allows two-stage detectors to be trained end-to-end. FPN\cite{lin2017feature} alleviates the scale mismatch between RPN's receptive fields and actual object size via feature pyramids. Cascade R-CNN\cite{cai2018cascade} applies a cascade architecture to regress BBoxes with a set of increasing IoU thresholds sequentially for progressive refinement. Mask R-CNN\cite{he2017mask} extends Faster R-CNN by constructing a proper mask branch that refines the detection results with the help of multi-task learning. On the other hand, one-stage detectors are popularized by YOLO\cite{redmon2016you} and SSD\cite{liu2016ssd} due to their computation efficiency. RetinaNet\cite{lin2017focal} with focal loss is proposed to address the extreme foreground-background class imbalance in dense object detection and gains higher accuracy than previous works. Other object detection methods focus on cascade procedures\cite{ouyang2017chained,gidaris2016attend,gidaris2015object,yang2016craft}, imbalance solution\cite{shrivastava2016training,singh2018sniper,ouyang2016factors,singh2018analysis,cao2019prime} and multi-scales adversarial mechanism\cite{zhu2019adapting}. They all make significant contribution to the current object detection field. 

\subsection{Imbalance in Object Detection}

When training an object detector, imbalance issue is a common but inevitable problem, which prevents well-designed model from being fully exploited. Solutions to alleviate this issue can be mainly divided into two categories so far. First is hard example mining method which relies on the hypothesis that hard examples are particularly significant to improve detection performance. OHEM\cite{shrivastava2016training} proposes a systematic approach considering the loss values of positive and negative samples and drives the focus towards hard examples according to their confidences. IoU-balanced sampling\cite{pang2019libra} is proposed to associate the hardness of examples with their IoUs and use a sampling method again for only negative examples rather than compute the loss function for the entire set. The second is the soft sampling method which scales the contribution of each example according to its corresponding importance to the training process. Focal loss\cite{lin2017focal} is the pioneer approach to dynamically assign more weights to the hard examples. GHM\cite{li2019gradient} defines the gradient density to handle disharmony of gradient norm distribution to avoid paying overmuch attention to outliers, which is shown useful for both classification and regression tasks. By alleviating imbalance in the training process, object detectors can be better trained, thus the better results can be obtained.

\section{Methodology}

In this section, we first illustrate the shortage of positive proposals in the begining of the training phase and describe R-CNN Gradient Annealing (RGA) strategy in detail. Next, we illustrate why the number of positive proposals is unbalanced across different testing samples. Finally, we introduce the structure of parallel R-CNN modules (PRM) and experimentally explain how PRM alleviates the problem mentioned above. 

\subsection{R-CNN Gradient Annealing} \label{sec:RGA}

\noindent \textbf{Shortage of Positive Proposals in the Beginning.} In the early training stage, since RPN cannot make accurate foreground/background discrimination, the large number of low-quality RPN generated proposals are fed into the following R-CNN module as training samples. Thus, the proposals can hardly satisfy the ``positive proposal definition'' (\emph{e.g.}, having an IoU which exceeds 0.5 with any ground-truth instance), resulting in a severe shortage of positive proposals in the early training stage. Although a fixed positive/negative sampling ratio, \emph{e.g.}, 1:3, is often used during training current detectors, a ``soft'' sampling ratio 1:3 is actually adopted in the practical implementation. For example, assuming batch size to be 512 for R-CNN, a sampling ratio 1:3 requires to sample 128 positive and 384 negative proposals from all the RPN proposals. In the early training stage, the number of all the positive ones $N_{real}^{+}$ provided by RPN is probably lower than 128. In this case, current implementations just use all the $N_{real}^{+}$ positive ones and $512-N_{real}^{+}$ negative ones to form the batch. So the actual positive/negative sampling ratio is smaller than 1:3, which is referred to ``soft'' sampling ratio 1:3 here. The insufficient positive proposals in the early training stage hurts the model performance, especially on the positive samples.

\begin{figure}[!t]
\begin{center}
\includegraphics[width=0.97\linewidth]{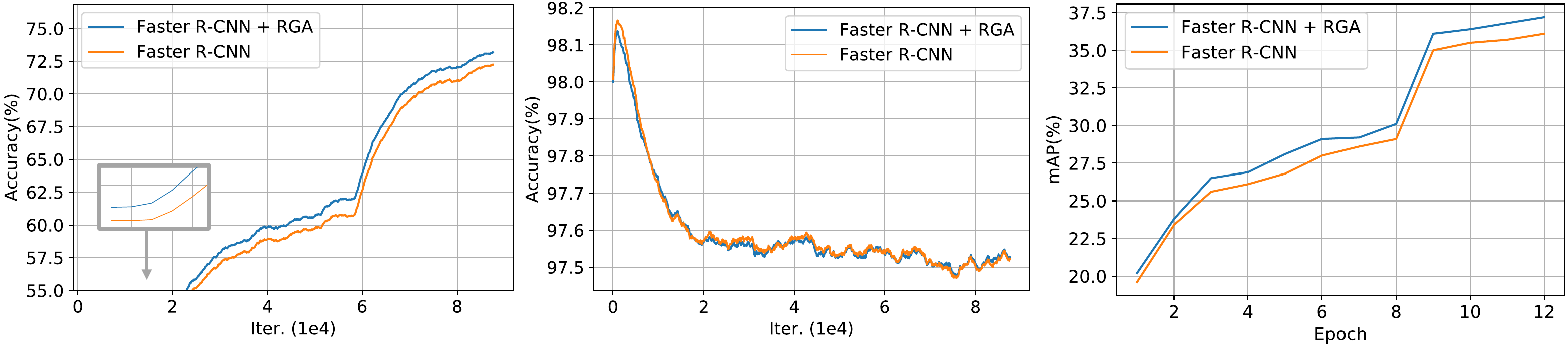}
\end{center}
   \caption{\textbf{Left:} Training accuracy of the positive proposals \emph{v.s.} the number of training iterations. \textbf{Middle:} Training accuracy of the negative proposals \emph{v.s.} the number of training iterations. \textbf{Right:} Validation performance \emph{v.s.} the number of training epochs.}
   \label{fprocess}
\end{figure}

For a better understanding, we analyze the model performance on both positive and negative samples, as the training progresses. From Fig.~\ref{fprocess} \textbf{Left} and Fig.~\ref{fprocess} \textbf{Middle}, one can observe that at the beginning of training, due to the overwhelming number of negative proposals, nearly all the proposals will be predicted as the negative. As the training progresses, the accuracy of positive proposals gradually increases while the accuracy of negative ones decreases until becoming stable. As can be seen, the entire training phase can be viewed as a process of learning what is positive. From this perspective, the shortage of positive training proposals in the beginning is no doubt an obstacle for training a high quality R-CNN module.

A straightforward countermeasure to the above issue is to copy the positive proposals multiple times, to achieve a ``hard sampling ratio''. However, ``hard sampling ratio'' will always cause performance drop according to our experiments (see in supplementary materials). The reason is when ``soft sampling ratio'' is applied, the R-CNN module is capable of modeling the intrinsic proportion of positive proposals \emph{v.s.} all proposals, which is beneficial to the testing performance, while ``hard sampling ratio'' conversely imposes a strong classification bias into R-CNN, which is proved to be harmful. Thus the potential solution to this issue needs to simultaneously raise the importance of positive proposals at early training stage, and leave the intrinsic proportion of positive proposals unchanged. 

\noindent \textbf{Magnifying and Annealing R-CNN Gradients.}  As we have mentioned in Sec.~\ref{sec:RGA} -- the entire training phase can be viewed as a process of learning what is positive. Fig.~\ref{fprocess} \textbf{Middle} shows that the classification accuracy of negative proposals first reaches 98\% then starts to drop gradually, while the classification accuracy of positive proposals stably increases. Such a phenomenon indicates that the gradients of positive proposals take over the training process at very early stages (\emph{i.e.} the average gradient of positive proposals is larger than negative proposals'). We thus propose to simultaneously magnify the gradients from both positive and negative proposals by a factor $\lambda$ because such an operation has a similar effect to only magnifying the weight of positive proposals while keeping the proportion of positive proposals \emph{v.s.} all proposals unchanged. Gradually decreasing the magnification factor $\lambda$ as the training progresses is introduced to guarantee that the gradients from both positive and negative proposals can always rival each other. We name such a solution ``R-CNN Gradient Annealing'' (RGA). A formal description of RGA is as follows.

\begin{center}
\vspace{-0.7cm}
\begin{equation}
\begin{split}
\theta _{t+1}=\theta_{t}-\alpha (\lambda \frac{\partial }{\partial \theta _{t}}{J}\left ( \theta_{t} \right )) \\
s.t.\quad  \lambda =\lambda_0 - \frac{(\lambda_0 -1)t}{T}\\
\end{split}
\end{equation}
\end{center}

where $\theta_t$ represents the parameters of R-CNN module in the $t$th optimization step, $\alpha $ is the current learning rate, $T$ is the total number of optimization steps, $\lambda_0$ is the initial magnification factor. $J$ is the loss function for R-CNN module.

\subsection{Parallel R-CNN Modules}

\noindent \textbf{Positive Proposal Imbalance in Testing Phase.} As aforementioned, a better per-image accuracy on the testing set can be achieved when the sample ratios on all the classes of the training set are more consistent with that of the testing set. Object detection can be viewed as proposal-level classification. Therefore, the positive/negative proposal ratio during training the R-CNN module is also required to be more consistent with that in the testing phase, for a better testing performance. Although the positive/negative proposal ratios in testing images are not directly decided by the ground-truth instance numbers, they are still highly correlated as shown in Fig.~\ref{fstart} (b). Therefore, we conclude that testing images with diverse ground-truth instance numbers require different positive/negative sampling ratios during training process of the R-CNN module, for the optimal testing results. Based on this finding, we claim that using a single positive/negative sampling ratio in R-CNN module training in the existing works is a sub-optimal solution, when facing the great diversity of ground-truth instance numbers of the testing set (\emph{e.g.}, MS COCO).

\begin{figure*}[!t]
\begin{center}
\includegraphics[width=0.95\linewidth]{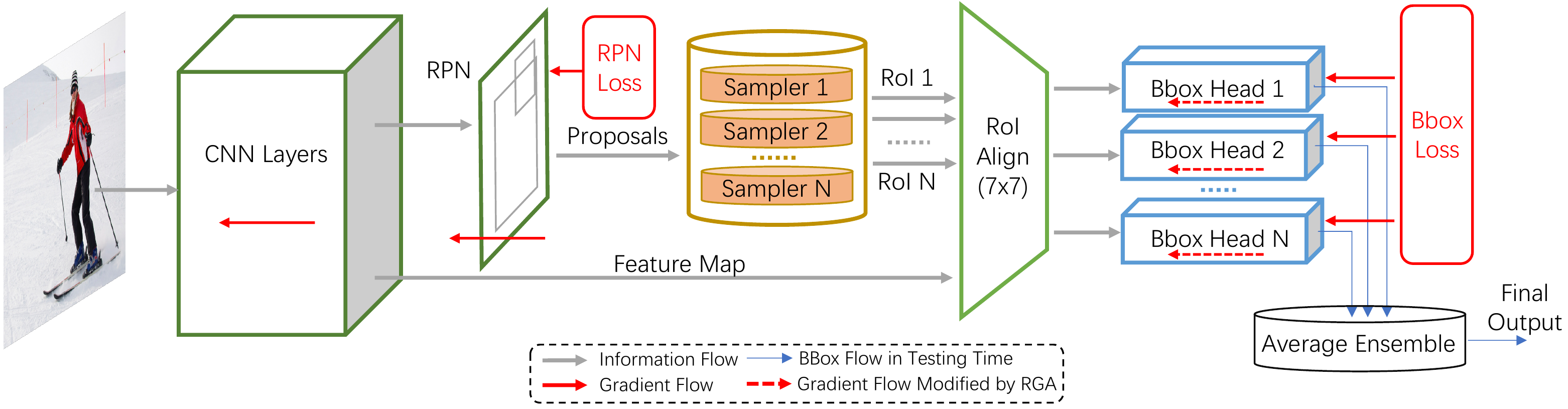}
\end{center}
   \vspace{-0.3cm}
   \caption{Model structure and information flow of our proposed Parallel R-CNN Modules and R-CNN Gradient Annealing strategy.}
   \vspace{-0.3cm}
   \label{fstructure}
\end{figure*}

\noindent \textbf{Model Structure.} To address the above issue, better consistency between the training sampling ratio and ground-truth instance number of each testing image is desired. To achieve this goal, an ideal solution could be: \textit{a Faster R-CNN with a set of parallel R-CNN modules trained with different positive/negative proposal sampling ratios; in the testing phase, the model dynamically dispatches each testing image to the best matched R-CNN module for prediction, based on its ground-truth instance number.} However, two obstacles impede the effective implementation of this solution. Firstly, the ground-truth instance number is unknown during testing, and the task of the accurate prediction of the ground-truth instance number remains difficult. Secondly, even if the accurate prediction of the ground-truth instance number is possible, the exact matching from each ground-truth instance number to the optimal training sampling ratio is also hard to decide.

To avoid the difficult per-image dispatch to the single optimal training sampling ratio, we propose to utilize the ``average ensemble'' strategy. Specifically, the model is a Faster R-CNN with a set of parallel R-CNN modules trained with different positive/negative proposal sampling ratios. In the testing phase, each R-CNN needs to process all the testing images. The final classification scores of the multiple R-CNN modules before softmax normalization are averaged to produce the final classification score. The bounding box regression output from the R-CNN module trained with the highest positive/negative sampling ratio is directly adopted. Notice that only positive proposals are used to train the BBox regression heads, and thus the training samples of other BBox regression heads are almost subsets of the one with the largest number of positive proposals. This is the reason why we fully trust the results from the BBox regression head trained with the most positive proposals. Fig.~\ref{fstructure} gives an illustration of Faster R-CNN with Parallel R-CNN Modules (PRM).

\subsection{Mechanism of PRM} 
In this section, we analyze the mechanism of PRM about how it benefits the detection on images with diverse ground-truth instance numbers. Specifically, two mechanisms are discovered, \emph{i.e.}, \emph{\textbf{Result Ensemble}} in the testing phase and \textbf{\emph{Gradient Ensemble}} in the training phase. We thus decouple the two mechanisms for further analysis.

\noindent \textbf{Result Ensemble.} In the testing phase, multiple R-CNN modules following a shared backbone and RPN allows ``Average Ensemble'' to be conveniently performed on each testing proposal. ``Average Ensemble'' effectively combines the decisions of the R-CNN modules biased to different class distributions. Therefore, the combined results are naturally better than that of an R-CNN module trained with a single sampling ratio, when encountering testing images with diverse ground-truth instance numbers.

\begin{figure*}[!t]
\begin{center}
\includegraphics[width=0.7\linewidth]{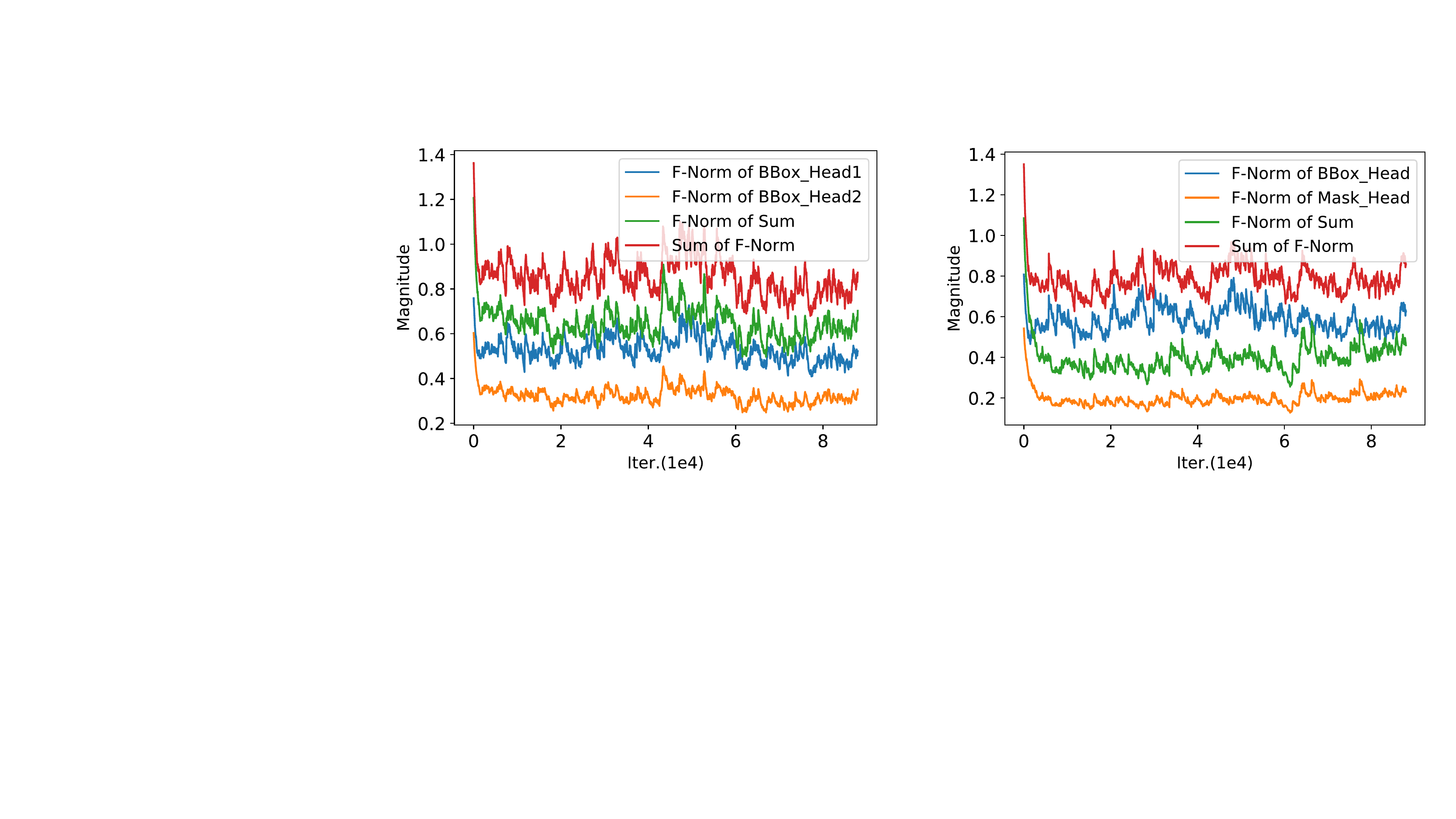}
\end{center}
   \vspace{-0.3cm}
   \caption{\textbf{Left} and \textbf{Right} are smoothed visualizations of the backward gradients on the weight of Block4.Conv1 in backbone, where F-Norm is the abbreviation of ``Frobenius Norm''.}
   \label{fgradients}
   \vspace{-0.5cm}
\end{figure*}

Notice that unlike image classification and semantic segmentation, ``average ensemble'' is not a common result ensemble technique in object detection. Due to the lack of clear correspondence of the detection boxes generated by different models, ``joint NMS'' which mixes the detection boxes together and then performs NMS among the mixed boxes is commonly utilized in the model ensemble in object detection. However, ``joint NMS'' has quadratic complexity w.r.t. the total number of detection boxes produced by all the models used for ensemble. In comparison, ``average ensemble'' in PRM introduces barely no extra costs, as only the multiple R-CNN modules are unshared, which usually contains only a few fully connected layers.

\noindent \textbf{Gradient Ensemble.} Apart from result ensemble in the testing phase, gradient ensemble in the training phase also plays a key role in PRM. Without result ensemble, PRM also enhances the detection performance of each individual R-CNN module compared to training them solely. The performance gain of each individual R-CNN module can only be attributed to the gradient ensemble in the shared backbone. This is similar to the phenomenon that the tasks have mutual benefit to each other in the multi-task learning. Gradients from different tasks are fused together in the shared backbone, strengthening the adaptation of the backbone to different tasks (\emph{i.e.}, testing images of diverse ground-truth instance numbers in our case). We plot the gradients of the weights of Block4.Conv1 in the shared backbone propagated from different R-CNN modules in Fig.~\ref{fgradients} \textbf{Left}. It can be seen in Fig.~\ref{fgradients} \textbf{Left} that the magnitude of gradients' vector sum from two R-CNN modules is always smaller than the sum of two gradients' magnitudes, which means the knowledge provided by two R-CNN modules are not exactly the same. As shown in Fig.~\ref{fgradients} \textbf{Right}, a similar phenomenon is observed in Mask R-CNN\cite{he2017mask} where the BBox head and the Mask head can be viewed as multi-task heads. 

\section{Experiments}

\subsection{Experiments on COCO} \label{sec4-1}

\noindent \textbf{Dataset.} COCO\cite{lin2014microsoft} is a widely used benchmark in object detection field. In this work, we train all our models on the COCO \textit{trainval35k} which consists of 118k images, and evaluate our results on the COCO $minival$ which consists of 5k testing images. Our evaluation metric follows the standard COCO style mean Average Precision (AP) at different BBox IoU thresholds. 

\begin{table}[!t]
\caption{Experimental results of our proposed RGA strategy and PRM across different backbones, learning rate schedules and other variants of two-stage detectors. Results are reported on COCO $minival$. The learning rate schedules of “1x”, “2x” follow the definitions in Detectron~\cite{girshick2018detectron}.} 
\centering
\begin{tabular}{lccllllll}
\hline
Method                       & Backbone          & Schedule & AP & AP$_{50}$ & AP$_{75}$ \\ \hline
Faster R-CNN\cite{ren2015faster}                 & ResNet-50         & 1x       &  36.1   & 58.1     &  38.8       \\
Faster R-CNN                 & ResNet-50         & 2x       &  37.3   & 59.0     &  40.5    \\
Faster R-CNN                 & ResNeXt-101\_32x4d & 1x       &  40.1   & 62.0     &  43.8    \\
Cascade R-CNN\cite{cai2018cascade}                & ResNet-50         & 1x       &  40.5   &  58.6    &   44.2    \\ \hline
Faster R-CNN w/ RGA                  & ResNet-50         & 1x       &  37.3   &  59.4    & 40.8      \\
Faster R-CNN w/ RGA              & ResNet-50         & 2x       &  38.4   & 59.9    & 41.8     \\
Faster R-CNN w/ RGA                  & ResNeXt-101\_32x4d & 1x       & 41.0    & 63.2     &  45.1    \\
Cascade R-CNN w/ RGA          & ResNet-50         & 1x       &  41.1  & 59.8      & 44.8   \\ \hline
Faster R-CNN w/ RGA+PRM         & ResNet-50         & 1x       &  38.1   & 60.0     &  41.6    \\
Faster R-CNN w/ RGA+PRM         & ResNet-50         & 2x       &  39.0   &  60.5    & 42.3  \\
Faster R-CNN w/ RGA+PRM         & ResNeXt-101\_32x4d & 1x       &  41.4   & \textbf{63.3}     &  45.4   \\
Cascade R-CNN w/ RGA+PRM & ResNet-50         & 1x       & \textbf{41.7}  & 60.0   &  \textbf{45.5} \\ \hline
\end{tabular}
\centering
\label{table4}
\end{table}

\noindent \textbf{Implementation Details.} To provide a strong baseline, we incorporate FPN\cite{lin2017feature} and RoI-Align\cite{he2017mask} into the naive Faster R-CNN\cite{ren2015faster}. If not specifically noted, in our paper, the term ``Faster R-CNN'' represents this modified version and optimized in ``jointly training'' manner. We train detectors on 8 GPUs (2 images per GPU) with an initial learning rate of 0.02, and decrease it by 0.1 at the 8th and 11th epoch. The magnification factor $\lambda$ is initialized to 7. Following the structure in \cite{lin2017feature}, our R-CNN module only contains 2 shared $fc$ and 2 separate $fc$ layers for classification and regression respectively. In our re-implemented Faster R-CNN\cite{ren2015faster}, the sampling ratio of positive/negative proposals is set to 1:3, which is the same as original papers. For PRM, we use two R-CNN modules with sampling ratios of 1:1 and 1:9. 

\noindent \textbf{Main Results.} Results on COCO \textit{minival} are presented in Table \ref{table4}. As shown in Table \ref{table4}, incorporating RGA into Faster R-CNN can bring 1.2\% improvement on AP. Such a gain reaches to 2.0\% after further adopting the structure of PRM (36.1\% \emph{v.s.} 38.1\%). RGA and PRM can still improve the baseline by 1.7\% (37.3\% \emph{v.s.} 39.0\%) when we increase the total training time from 1x to 2x to make sure the models are better optimized. Furthermore, we evaluate our method on a better backbone -- ResNeXt-101 and a better two-stage detector -- Cascade R-CNN. Results in Table \ref{table4} tell that our proposed RGA strategy and PRM can yield consistent improvements across different learning rate schedules, backbones and variants of two-stage detectors. It is worth mentioning that our method can especially improve the performance under AP$_{75}$ by 2.8\% (38.8\% \emph{v.s.} 41.6\%) which illustrates the effectiveness of RGA and PRM under more severe evaluation protocol.


To better compare our method with other state-of-the-art detectors, we evaluate our method on COCO \textit{test-dev}. As shown in Table \ref{sota}, a single Faster R-CNN with RGA and PRM with the backbone ResNeXt-101 can reach 42.9\% on AP, which is comparable to Libra R-CNN. Our method can also improve the performance of Cascade R-CNN by 1.0\%. After incorporating the cascade mechanism~\cite{cai2018cascade} into our model, the performance of a single model can reach 45.3\% without any bells and whistles (\emph{e.g.}, longer learning rate schedule, deformable convolution and training/testing time augmentation).

\begin{table}[!t]
\caption{Comparisons with state-of-the-art methods on COCO \textit{test-dev}. * means our re-implemented results. All the two-stage detectors follow 1x training schedule. ResNeXt-101 means ResNeXt-101\_64x4d by default.} 
\centering
\begin{tabular}{lcllllll}
\hline
Method                       & Backbone          & AP & AP$_{50}$ & AP$_{75}$ & AP$_S$ & AP$_M$ & AP$_L$ \\ \hline
YOLOv2\cite{redmon2017yolo9000}                & DarkNet-19           &  21.6   & 44.0     &  19.2    &  5.0   &  22.4 & 35.5    \\
SSD512\cite{liu2016ssd}                 & ResNet-101        &  31.2   & 50.4     &  33.3    &  10.2  & 34.5   &  49.8   \\
RetinaNet\cite{lin2017focal}                 & ResNet-101-FPN    &  39.1   & 59.1     &  42.3    &  21.8   & 42.7    &  50.2   \\ \hline
Faster R-CNN\cite{ren2015faster} & ResNet-101-FPN          &  36.2   &  59.1    &   39.0   &  18.2   & 39.0    &  48.2   \\
Mask R-CNN\cite{he2017mask}                 & ResNet-101-FPN       &  38.2   &  60.3    & 41.7     & 20.1   & 41.1   & 50.2    \\ 
Libra R-CNN\cite{pang2019libra}                  & ResNeXt-101-FPN        & 43.0    & 64.0     &  47.0    & 25.3    & 45.6    & 54.6    \\ \hline
Faster R-CNN*              & ResNet-101-FPN          &  38.7  & 60.8    & 42.3    & 22.3   &  42.2    & 48.5    \\
Cascade R-CNN*              & ResNet-101-FPN          &  42.1   &  61.0   &  46.0    & 23.5   & 45.5  &  54.7  \\ \hline
Faster R-CNN w/ RGA+PRM         & ResNet-101-FPN          &  40.2   & 62.1     & 43.9     &  23.4   & 43.6    & 50.7    \\
Faster R-CNN w/ RGA+PRM        & ResNeXt-101-FPN    &  42.9  & \textbf{64.9}  & 46.7 & 25.6   &   46.2  & 53.7 \\
Cascade R-CNN w/ RGA+PRM         & ResNet-101-FPN       &  43.1   & 61.5     &  47.0      &  24.1   & 46.1    & 55.4    \\
Cascade R-CNN w/ RGA+PRM & ResNeXt-101-FPN          & \textbf{45.3}  & 64.0   & \textbf{49.5}  & \textbf{26.6} & \textbf{48.1}  & \textbf{57.8}  \\ \hline
\end{tabular}
\centering
\label{sota}
\vspace{-0.2cm}
\end{table}

\subsection{Ablation Study} \label{sec:ablation}

\noindent \textbf{R-CNN Gradient Annealing.} We study how the annealing of magnification factor $\lambda$ helps improve the final prediction results by setting $\lambda$ as a constant value. Comparing the 1st, 2nd and 5th line in Table \ref{table5}, we find that although magnifying the gradients in R-CNN module 7 times can bring 0.6\% improvement on AP compared to its baseline (36.1\%), with the introduction of annealing, the final performance can be further improved by 0.6\%, which proves the effectiveness of gradient annealing. Since a new hyper-parameter $\lambda$ is introduced, we test the influence of different $\lambda_0$ in Table \ref{table5}. Results indicate that our method is not sensitive to the value of $\lambda$ within the range of $[5,9]$. It deserves to be mentioned that RGA brings no extra computation cost during testing time. All of these merits make our RGA strategy more applicable.

\begin{table}[htbp]
\caption{Varying $\lambda_0$ in RGA. The RGA strategy yields consistent improvement over baseline when $\lambda_0$ varies from 5 to 9.}
\centering
\begin{tabular}{cc|ccc}
\hline
$\lambda_0$ & Anneal & AP & AP$_{50}$ & AP$_{75}$ \\ \hline
1           & -      &  36.1  &     58.1 &   38.8   \\
7           & -      &  36.7  &     58.5 &   39.8   \\ \hline
3           & \checkmark       & 36.9   & 59.0     &  40.2  \\
5           & \checkmark       & 37.2   & 59.0     &  \textbf{40.8}    \\
7           & \checkmark       & \textbf{37.3}   & \textbf{59.4}     &  40.6  \\
9           & \checkmark       & 37.2   & 59.3     &  40.0  \\ \hline
\end{tabular}
\centering
\label{table5}
\end{table}

\noindent \textbf{Sampling Ratios.} Sampling ratio is crucial to PRM. To understand this issue better, we investigate the case of two R-CNN modules and run experiments with a set of different sampling ratio pairs. Results are presented in Table \ref{table2}. It shows that when the two sampling ratios are 1:1 and 1:9, the detector achieves the highest AP -- 37.3\%. When the two sampling ratios are 1:1 and 1:1, the detector achieves the worst AP which is 36.8\%. According to such results, we can conclude that the detection performance is correlated to the gap between the two chosen sampling ratios in a pair. Specifically, the larger gap between two sampling ratios is, the better final result will be achieved. In addition, to prove the performance gain of PRM does not come from more parameters, we train two separate Faster R-CNN models with the sampling ratios of 1:1 and 1:9, whose number of parameters is far larger than a single Faster R-CNN with PRM. The results are presented in the second row in Table~\ref{table2}. Comparing the results in the second and fifth rows, we can see that the performance of Faster R-CNN with PRM exceeds the ensemble of two separate Faster R-CNN models. It means the benefit brought by Gradient Ensemble is larger than the number of parameters, which is one main source of gain of PRM.

\begin{table}[h]
\caption{Varying sampling ratios across R-CNN modules. ``R'' and ``E'' are the abbreviations of ``R-CNN'' and ``Ensemble'', respectively. ``Faster R-CNNx2'' stands for two separate Faster R-CNN models.}
\centering
\begin{tabular}{l|cc|ccc}
\hline
\multirow{2}{*}{method}      & \multicolumn{2}{c|}{sample ratios} & \multicolumn{3}{c}{AP}                 \\ \cline{2-6} 
                             & R1             & R2             & R1   & \multicolumn{1}{c|}{R2}   & E    \\ \hline
Faster R-CNN                 & 1:1            & -              & 36.3     &
                    \multicolumn{1}{c|}{-}          & - \\ \hline
Faster R-CNNx2                 & 1:1            & 1:9              & 36.3     & 
                    \multicolumn{1}{c|}{34.9}          & 36.5 \\ \hline
\multirow{3}{*}{Faster R-CNN w/ PRM} & 1:1            & 1:1            &   36.7   &       \multicolumn{1}{c|}{36.7}     &   36.8   \\
                             & 1:1            & 1:5            &   36.8   & \multicolumn{1}{c|}{36.9}     &   37.1   \\
                             & 1:1            & 1:9            & 37.0 & \multicolumn{1}{c|}{36.9} & \textbf{37.3} \\ \hline
\end{tabular}
\centering
\label{table2}
\end{table}


\begin{table}[h]
\caption{With the third R-CNN module of the sampling ratio 1:3, the performance drops 0.2\% both with and without RGA.}
\centering
\begin{tabular}{cc|ccc}
\hline
    3rd R-CNN  & w/ RGA & AP & AP$_{50}$ & AP$_{75}$ \\ \hline
        -        &    -      &  37.3  &  58.8    &40.4      \\
        \checkmark    &   -       &  37.1  &  58.8  & 40.1     \\ \hline
        -        &    \checkmark  &  38.1  &  60.0    &41.6      \\
        \checkmark    &    \checkmark  &  37.9  &   59.6   & 41.2 \\ \hline
\end{tabular}
\centering
\label{table_num_heads}
\end{table}

\noindent \textbf{Number of R-CNN Modules.} The number of R-CNN modules is another hyper-parameter in PRM. Following the conclusion we draw in the previous paragraph, we choose PRM with two R-CNN modules, with the sampling ratios of which are 1:1 and 1:9 as our baseline. After adding the third R-CNN module with the sampling ratio of 1:3, as shown in Table \ref{table_num_heads}, we observe a little performance drop both with and without RGA (-0.2\%), which indicates that two R-CNN modules are sufficient. Such phenomenon probably results from that while all R-CNN modules are put on one shared backbone, as the number of R-CNN modules increases, jointly optimizing them may cause over-amplification of the gradients from different losses in the backbone.

\subsection{Further Analysis}

\noindent \textbf{The Gain of RGA.} The performance gain of RGA has been shown in Sec.~\ref{sec4-1}. However, where does such gain come from still remains unverified. The answer is hidden in Fig.~\ref{fprocess} which describes two full training processes for a baseline Faster R-CNN and Faster R-CNN with RGA, respectively. Fig.~\ref{fprocess} \textbf{Middle} shows that the training accuracy for negative proposals are nearly identical when with and without RGA, which means RGA can not improve model's ability of identifying negative proposals. However, Fig.~\ref{fprocess} \textbf{Left} shows that after applying RGA, the training accuracy for positive proposals is consistently better than baseline. Such results verify our assumption that by magnifying the gradients from both positive and negative proposals, it is positive proposals whose impact are amplified so that the R-CNN module can better identify positive proposals. Consequently, the validation performance is improved.

\begin{figure}[h]
\begin{center}
\includegraphics[width=0.7\linewidth]{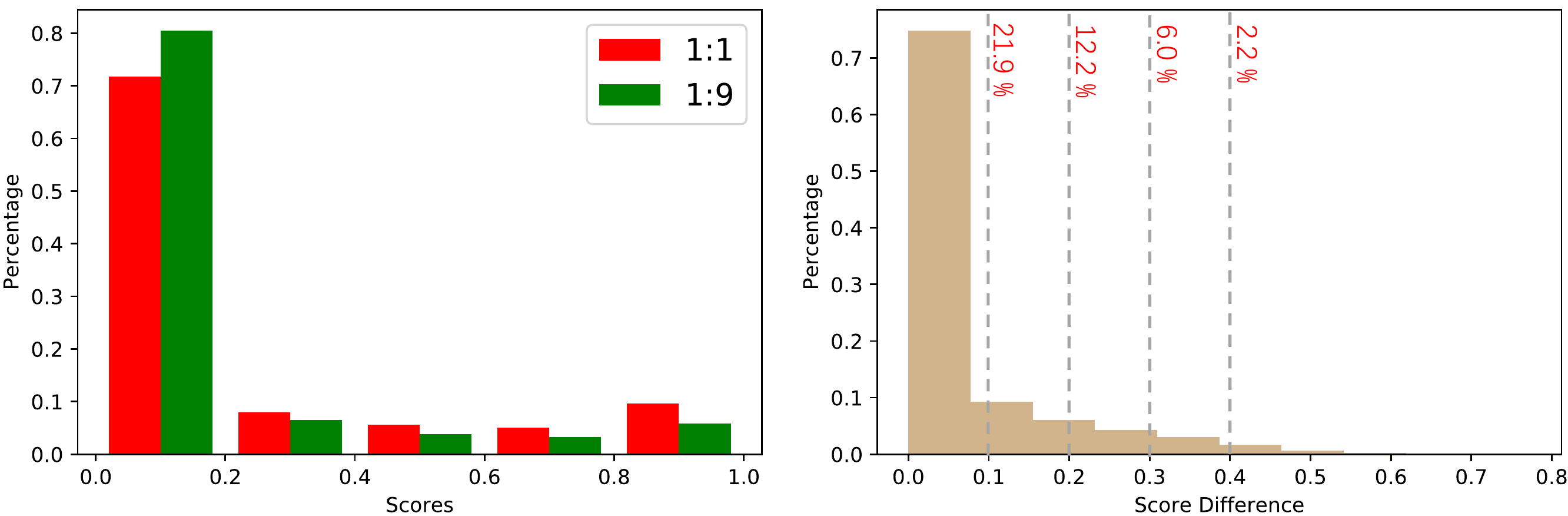}
\end{center}
\vspace{-0.3cm}
   \caption{Difference between two R-CNN modules. \textbf{Left:} distribution of predicted scores. \textbf{Right:} distribution of the absolute difference of predicted scores between two R-CNN modules.}
   \label{fdistinguish}
   \vspace{-0.3cm}
\end{figure}

\noindent \textbf{Comparisons between Different R-CNN Modules.} Considering the same optimization of objectives used in different R-CNN modules, it is natural to ask this question: is it possible that those different R-CNN modules actually learn similar parameters to each other? We try to answer this question from the discrepancy between their prediction scores. The detector we use is a well-trained Faster R-CNN with two R-CNN modules trained with the sampling ratios of 1:1 and 1:9. Fig. \ref{fdistinguish} \textbf{Left} shows the distribution of predicted scores on COCO $minival$ from two R-CNN modules. From Fig. \ref{fdistinguish}, we can learn that the R-CNN module trained with the smaller sampling ratio of positive proposals tends to predict lower scores, and vice versa, which is already stated in Sec~\ref{sec:intro}. Fig. \ref{fdistinguish} \textbf{Right} visualizes the absolute difference of scores between corresponding outputs. It shows that there are more than 21.9\% pairs of prediction scores have the absolute difference larger than 0.1. Such a phenomenon verifies our claim in Sec.~\ref{sec:intro} that R-CNN modules with different sampling ratios can show different prediction biases, which is the reason why output ensemble can bring improvements. \textbf{The improvements of PRM on different subset of COCO $minival$ can be seen in Fig.~\ref{fdivide}}.

\subsection{Experiments on CrowdHuman}
To prove the generalization ability of RGA and PBM, we evaluate them on an extra dataset -- CrowdHuman\cite{shao2018crowdhuman}. CrowdHuman is a benchmark for detecting human body in the crowded situation. It contains $15,000$, $4,370$, and $5,000$ images for training, validation and testing, respectively. On average, there are around 23 persons per-image, making CrowdHuman a challenging benchmark. In CrowdHuman, there are three kinds of annotations: \textit{full body}, \textit{visible body} and \textit{head}. We focus on \textit{full body} in our experiments. All the configurations follow the original paper\cite{shao2018crowdhuman}. When applying our method, we set $\lambda_0$ to 7 and use Faster R-CNN with PRM trained with the sampling ratios of 1:1 and 1:9. The log-average-missing-rate (mMR, lower is better) and AP$_{50}$ are reported in Table \ref{table7}. As can be seen in Table \ref{table7}, our RGA and PRM bring a remarkable reduction of 2.29\% on mMR and 1.35\% improvement on AP$_{50}$, which proves the effectiveness of our method across various detection tasks.

\begin{table}[!h]
\vspace{-0.3cm}
\caption{RGA and PRM bring remarkable improvement on CrowdHuman dataset. * stands for our re-implementation result.}
\centering
\begin{tabular}{l|c|c}
\hline
method              & mMR & AP$_{50}$ \\ \hline
Faster R-CNN Baseline in~\cite{shao2018crowdhuman}        &  50.42   &   84.95    \\ \hline
Faster R-CNN*  &  47.42   &  85.02     \\
Faster R-CNN w/ PRM &  46.30   & 85.43  \\
Faster R-CNN w/ RGA+PRM  &  \textbf{45.13}  &  \textbf{86.37}    \\ \hline
\end{tabular}
\centering
\label{table7}
\vspace{-0.5cm}
\end{table}

\section{Conclusion} 

In this paper, we propose R-CNN Gradient Annealing strategy, a gradient manipulation operation to alleviate the imbalance of the number of positive proposals in the training phase. We also propose a new design of two-stage object detector PRM which deploys several parallel R-CNN modules trained with different positive/negative proposal sampling ratios on a same backbone. Such design overcomes the imbalance of positive proposals across testing images. These two innovations can totally brings 2.0\% improvement based on a modified Faster R-CNN baseline, which strongly validates the utility of the proposed approach.

\section{Appendix} 
\appendix

\section{Hard Sampling and Soft Sampling}
\vspace{-1.2cm}
\begin{table}[!h]
\caption{Performance comparison between hard and soft sampling strategies.}
\centering
\begin{tabular}{c|cc}
\hline
sample ratio & hard method & soft method \\ 
\hline
1:1 & 33.3 & 36.3\\
1:3 & 35.2 & 36.1\\ 
1:5 & 35.5 & 35.7\\ 
1:7 & 35.2 & 35.4\\
\hline
\end{tabular}
\centering
\label{table10}
\vspace{-0.2cm}
\end{table}

As we stated in Sec. 1 in our submission paper, in real experiments, the number of positive proposals can hardly meet the desired number given the sampling ratio of 1:1 or 1:3. Copying the positive proposals multiple times to achieve a ``hard sampling ratio'' is a more natural solution to enhance the gradients from positive proposals than RGA. However, the performances of ``hard sampling'' are consistently worse than ``hard sampling'' because ``hard sampling ratio'' imposes a strong classification bias into R-CNN, which could be harmful. Another key difference between ``hard sampling ratio'' and RGA is that ``hard sampling ratio'' needs to reduce the number of negative proposals. In that way, the diversity of negative proposals will be reduced while RGA can avoid that. That is why RGA can yield better results while ``hard sampling ratio'' can not although they both enhance the gradients of positive proposals.  Experimental results can be seen in Table \ref{table10}.

\clearpage

\vspace{-0.2cm}
\section{Implementation of RGA}
\vspace{-0.2cm}
RGA is easy to implement. Here is our implementation of R-CNN Gradient Annealing strategy in MMDetection. 
\begin{lstlisting}
#mmdetection/mmdet/core/utils/dist_utils.py
def after_train_iter(self, runner):
    runner.optimizer.zero_grad()
    runner.outputs['loss'].backward()
    
    #####################
    # RGA, alpha0=7
    #####################
    weight = (7. - 6. * runner.iter / runner.max_iters)
    for name, param in runner.model.module.named_parameters():
        if 'bbox_head' in name.split('.')[0]:
            param.grad *= weight
    #####################
    
    allreduce_grads(runner.model.parameters(), self.coalesce,
                    self.bucket_size_mb)
    if self.grad_clip is not None:
        self.clip_grads(runner.model.parameters())
    runner.optimizer.step()
\end{lstlisting}

%
%
\bibliographystyle{splncs04}
\bibliography{egbib}

\end{document}